\documentclass[default]{sn-jnl}

\usepackage{geometry}
\usepackage{graphicx}
\usepackage{multirow}
\usepackage{amsmath,amssymb,amsfonts}
\usepackage{amsthm}
\usepackage{mathrsfs}
\usepackage[title]{appendix}
\usepackage{xcolor}
\usepackage{textcomp}
\usepackage{manyfoot}
\usepackage{booktabs}
\usepackage{algorithm}
\usepackage{algorithmicx}
\usepackage{algpseudocode}
\usepackage{listings}
\usepackage{inputenc}
\usepackage{hyperref}
\usepackage{longtable}
\usepackage[section]{placeins}
\usepackage[justification=centering]{caption}
\usepackage{lscape}
\usepackage[none]{hyphenat}
\hypersetup{pdfauthor={Name}}
\usepackage{bbding}
\usepackage{pifont}
\usepackage{wasysym}
\newcommand{\xmark}{\ding{55}}

\begin{document}

\title[Exploring Machine Learning Models for Federated Learning: A Review of Approaches, Performance, and Limitations]{Exploring Machine Learning Models for Federated Learning: A Review of Approaches, Performance, and Limitations}
\author*[1]{\fnm{Elaheh} \sur{Jafarigol}}\email{elaheh.jafarigol@ou.edu}
\author[2]{\fnm{Theodore} B. \sur{Trafalis}}
\author[2]{\fnm{Talayeh} \sur{Razzaghi}}
\author[3]{\fnm{Mona}\sur{Zamankhani}}
\affil*[1]{\orgdiv{Data Science and Analytics Institute}, \orgname{University of Oklahoma}, \orgaddress{\street{202 W. Boyd St., Room 409}, \city{Norman}, \postcode{73019}, \state{Ok}, \country{USA}}}
\affil[2]{\orgdiv{School of Industrial and Systems Engineering}, \orgname{University of Oklahoma}, \orgaddress{\street{202 W Boyd St., Room 124}, \city{Norman}, \postcode{73019}, \state{OK}, \country{USA}}}

\affil[3]{\orgdiv{Department of Industrial Engineering}, \orgname{Isfahan University of Technology}, \orgaddress{\city{Isfahan}, \country{Iran}}}

\abstract{In the growing world of artificial intelligence, federated learning is a distributed learning framework enhanced to preserve the privacy of individuals' data. Federated learning lays the groundwork for collaborative research in areas where the data is sensitive. Federated learning has several implications for real-world problems. In times of crisis, when real-time decision-making is critical, federated learning allows multiple entities to work collectively without sharing sensitive data. This distributed approach enables us to leverage information from multiple sources and gain more diverse insights.
This paper is a systematic review of the literature on privacy-preserving machine learning in the last few years based on the Preferred Reporting Items for Systematic Reviews and Meta-Analyses (PRISMA) guidelines. Specifically, we have presented an extensive review of supervised/unsupervised machine learning algorithms, ensemble methods, meta-heuristic approaches, blockchain technology, and reinforcement learning used in the framework of federated learning, in addition to an overview of federated learning applications.  This paper reviews the literature on the components of federated learning and its applications in the last few years. The main purpose of this work is to provide researchers and practitioners with a comprehensive overview of federated learning from the machine learning point of view. A discussion of some open problems and future research directions in federated learning is also provided.}


\maketitle

\section{Introduction}\label{sec1}
\begin{figure}[!hbt]
\centering
\includegraphics[width = 1\textwidth, keepaspectratio]{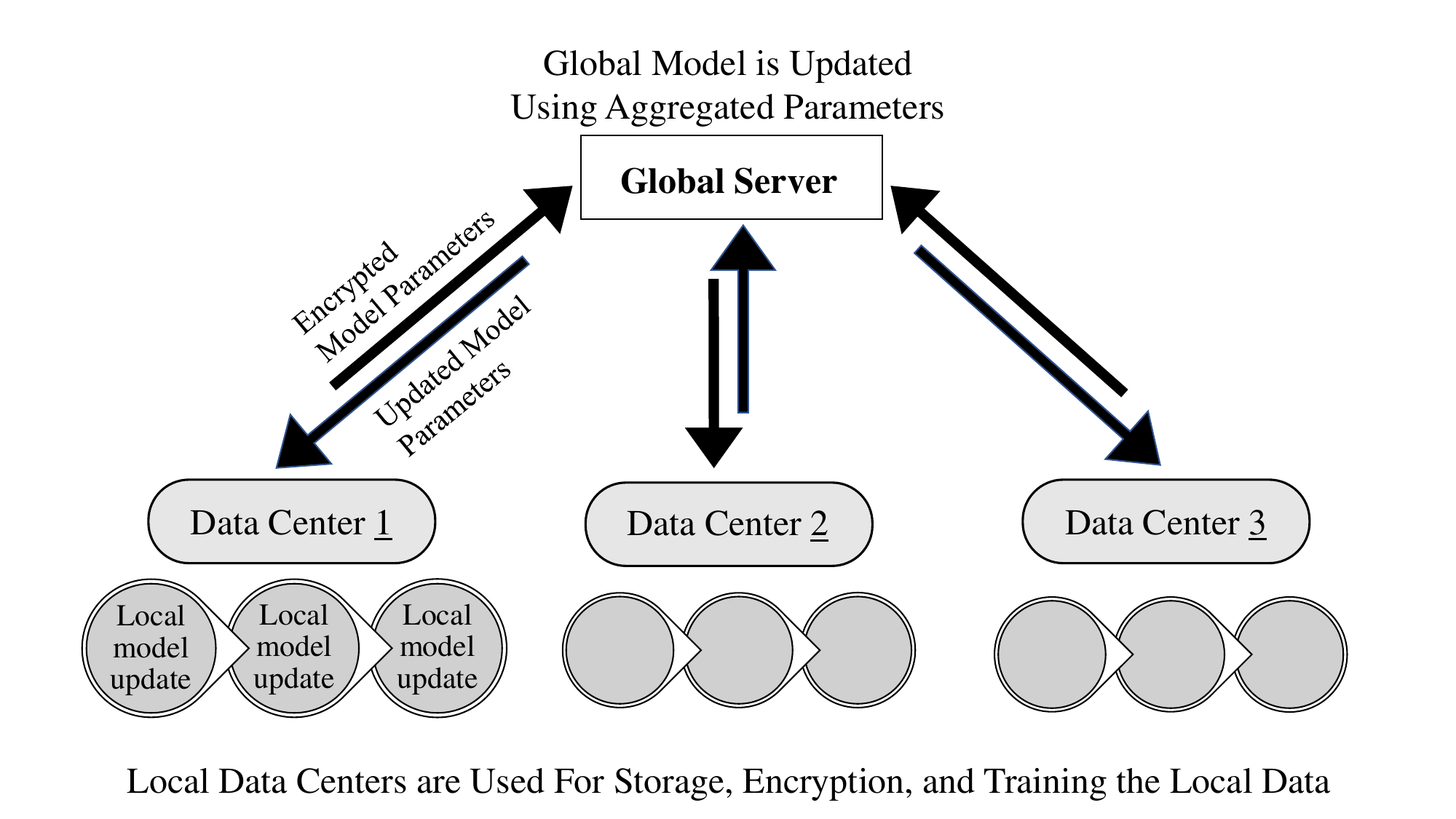}
\caption{The Federated Learning Framework \cite{mcmahan2017communication}}
\label{framework}
\end{figure}

Privacy is the individuals' right to control their personal information. With the advances in data-driven technologies, ensuring privacy protection has become more challenging. Privacy and federated learning are intertwined concepts in machine learning. Privacy and federated learning are intertwined concepts in machine learning. Federated learning offers a promising solution by enabling collaborative model training without exposing sensitive data. This decentralized approach preserves privacy by design, allowing organizations and individuals to collaborate while minimizing the risk of data breaches or unauthorized access. With privacy at its core, federated learning has become a possible solution for scenarios where the data is sensitive and individual privacy is a concern. An algorithm is considered private if the outcome of the analysis is the same, whether any arbitrary individual data is part of the dataset or not \cite{dwork2004privacy}. This definition is the basis for privacy protection mechanisms in federated learning. 
Federated learning is a broad term that includes different aspects of data collection, storage, analysis, and communication in a decentralized information system where data centers can not disclose data for learning purposes. The term federated learning was introduced in the paper published based on the results of a research project carried out at Google in 2016 for text input prediction on mobile devices \cite{mcmahan2017communication}. The authors designed a collaborative environment for a group of devices referred to as clients, coordinated by a central server, also known as a service provider. They also conducted an empirical study on the model using different model structures on benchmark datasets for image classification and language processing applications.\newline 
The ideas behind federated learning have been around for decades, but thanks to the abundant data that is available to us and advances in computation power, we are able to efficiently train machine learning models on a network of decentralized data sources. Advances in the field of machine learning, especially deep learning, allow us to build powerful models that can accurately analyze different data types and provide valuable insights. Without the need to collect data in one place, we can leverage multiple sources of data, knowing that individuals' privacy is preserved while reducing the costs of data storage and achieving high-quality results. Distributed machine learning algorithms create an environment where data storage and training happen on the group of distributed machines. However, the main difference between distributed learning and federated learning is the notion of privacy, and the advances in data analysis capacities made over the last decade allow us to develop and deploy privacy-preserving algorithms at scale \cite{li2018privacy, team2017learning, tan2022federated}.
\newline 
Federated learning is the process of training data locally and improving the global model. In the federated learning framework, as shown in Figure \ref{framework}, the data is stored in local data centers, and limited information required for the learning task is privately communicated with the central server. 
This architecture is called a client-server design. If the data centers are both responsible for the task of storage and aggregation, the architecture is called peer-to-peer. Suppose $n$ is the total number of sample points. In that case, $K$ is the total number of clients, $n_k$ is the total number of sample points on client $k$, and $\eta$ is the learning rate. The goal of federated learning is to minimize the objective function $f$, also known as the loss function, where $f_{k}(\delta)$ is the loss function, and $\delta$ is the evaluation value for the $\textit{k-th}$ client: 
\begin{equation}
    f(\delta) := \displaystyle\sum_{k=1}^{K} \frac{n_k}{n}f_k(\delta)
\end{equation}
In this equation, $\delta$ is updated after each iteration until we reach the optimal solution or the number of iterations set as the stopping criterion is satisfied. This optimization problem is solved using a federated Stochastic Gradient Descent (SGD) method, which is described in algorithm \ref{SGD_algo}. This algorithm shows that the gradient steps are taken by each client, and the model parameters are calculated as $\delta_{t+1} \leftarrow \delta_t - \eta g_k$.
\begin{algorithm}
\caption{Federated Stochastic Gradient Descent}\label{SGD_algo}
\begin{algorithmic}
\Require $\delta_t, \nabla f(\delta_t), n, n_k, \eta$
\Ensure $C<1$ 
\newline\Comment{a subset of clients is selected at each round}
\State  $\delta_t$:= the current state of the evaluation value
\While{$f(\delta_t)$ $\neq$ optimal}
\State  $g_k = \nabla f_k(\delta_t)$ 
\newline\Comment{$g_k$ := the gradient of client k}
\State $\delta_{t+l} \leftarrow \delta_t - \eta\nabla f(\delta_t) = \delta_t - \sum_{k=1}^{K} \frac{n_k}{n}g_k$ \newline\Comment{Aggregation of client gradients to create a new model at the central server}
\EndWhile
\end{algorithmic}
\end{algorithm}
\newline 
\newline
Federated learning affects the modeling step of the Cross-Industry Standard Process for Data Mining (CRISP-DM), which starts from local data storage in data centers to communicate with the central server to iteratively aggregate the model parameters and update the global model to achieve the desired learning accuracy in data centers. 
\begin{figure}[!h]
\centering
\includegraphics[width = 0.95\textwidth, keepaspectratio]{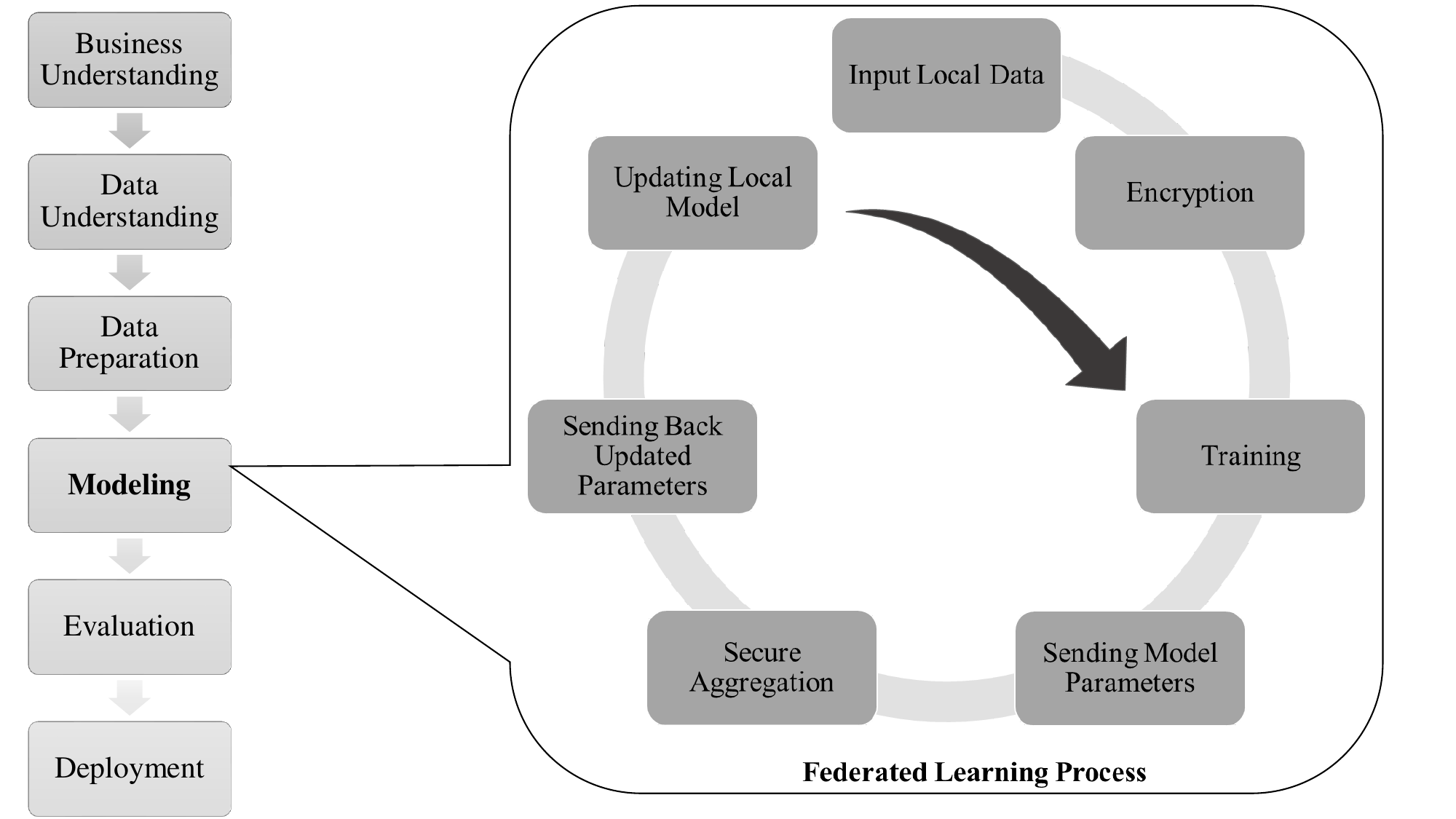}
\caption{Data Mining Process in Federated Learning Based on CRISP-DM Model}
\label{Cycle}
\end{figure}
Figure \ref{Cycle} shows the Federated learning life cycle embedded in CRISP-DM. 
\section{Applications}
Federated learning has been applied to various fields when the data is sensitive and scattered across multiple servers or devices. In particular, federated learning is a compelling approach in healthcare, IoT, and crisis management, where information limitation is an issue.
\subsection*{IoT} 
Industry 4.0 is the era of interconnected physical and digital technologies. Through the fourth industrial revolution, smart operations evolved, and the demand for informed and data-driven solutions increased. In the digital world, data is constantly generated in texts, images, and measurements from thousands of sensors and devices, which require powerful systems to perform extensive computations for data processing. Smart cities, cloud-based technologies, edge computing, and IoT need reliable, secure, real-time analysis tools. This has increased the demand for systems that can address scalability, interoperability, resource limitations, and privacy issues \cite{lim2020federated, khan2020IoT}. 

Nguyen et al. \cite{Nguyen2021survey} surveyed the application of federated learning to leverage the data on IoT devices for smart cities and industries, leading to advances in healthcare, transportation, and Unmanned Aerial Vehicles.  Khan et al. \cite{Khan2021survey} surveyed different aspects of federated learning for IoT applications. The authors compared and evaluated the methods from robustness, quantization, sparsification, scalability, security, and privacy perspectives. Varlamis et al. \cite{varlamis2022using} used federated learning to find energy-saving solutions based on sensor data. 
Federated learning offers several benefits when applied to Internet of Things (IoT) applications such as:
\begin{itemize}
    \item Privacy preservation: This is the primary advantage of federated learning. IoT devices often collect sensitive personal information, such as health data or home automation preferences, and federated learning allows these devices to train machine learning models without exposing individuals to possible harm and misuse of data. 
    \item Reduced data transfer: IoT devices are often resource-constrained, making it inefficient and costly to transmit large volumes of data to a central server. By keeping data on the edge, federated learning reduces the need for extensive data transfer and lowers communication overhead.
    \item Edge computing: Federated learning fits well with edge computing, where data processing occurs closer to the source (IoT devices) rather than in a data center. Therefore, minimizing latency and improving real-time decision-making which is crucial for applications like autonomous vehicles and industrial automation.
\end{itemize}
Some of the challenges and limitations of federated learning in this domain are:
\begin{itemize}
    \item Heterogeneity: IoT devices come in various shapes, sizes, and capabilities, making them challenging to harmonize for federated learning. Ensuring that models can be trained effectively across diverse IoT ecosystems is a significant challenge. In federated learning, some devices may also have more data or contribute more frequently to model updates than others. Managing the data across IoT devices can affect the fairness and accuracy of the learned models.
    \item Communication overhead: Aggregating model updates from numerous IoT devices can be challenging, especially when dealing with intermittent connectivity, device failures, or adversarial behavior. Robust aggregation methods are needed to handle these scenarios.
    \item Computational overhead: Federated learning can be computationally intensive, which may be problematic for resource-constrained IoT devices. Balancing model training with energy efficiency and processing power is a limitation that needs to be addressed.
    \item Scalability: As the number of IoT devices increases, managing federated learning across the network becomes more challenging.
\end{itemize}
Federated learning holds great promise for IoT applications. However, addressing challenges and overcoming limitations is essential to fully harness the potential of federated learning in the rapidly expanding IoT ecosystem.
\subsection*{Healthcare} 
Federated learning is a promising approach for learning from healthcare data, which is highly regulated and cannot be openly shared with the public. Therefore, if privacy is ensured, it can significantly benefit from utilizing artificial intelligence and machine learning to move towards personalized healthcare and computer-aided diagnosis. Federated learning creates a global model of decentralized data, such as the data from hospitals, labs, and clinical trials without direct access to the data \cite{xu2021federated}. For instance, Feki et al. \cite{feki2021federated} utilized a federated learning approach to build a powerful model to classify COVID-19 X-rays based on data collected from multiple institutes. Rieke et al. \cite{rieke2020future} explored the existing literature on federated learning for healthcare with the challenges and open problems in digital healthcare. 
Some of the benefits of federated learning in healthcare applications are:
\begin{itemize}
    \item Privacy preservation: Healthcare data is highly sensitive and subject to strict privacy regulations. Federated learning enables healthcare institutions to collaborate on model training without sharing raw patient data, preserving privacy and compliance with regulations like HIPAA.
    \item Large-scale data utilization: Without privacy concerns, healthcare organizations can tap into a vast pool of data from various sources, including hospitals, clinics, wearable devices, and electronic health records, to advance machine learning and computer-aided diagnosis.
    \item Personalized medicine: by leveraging patient-specific data, personalized treatment plans become more viable, leading to more effective and tailored healthcare interventions.
\end{itemize}
However, some of the challenges and limitations are:
\begin{itemize}
    \item Heterogeneity: Health data comes from various hospitals, clinics, personal health monitoring devices, and more. Differences in data formats, quality, completeness, and availability pose challenges for integrating the data and achieving accurate and reliable results.    
    \item Regulatory compliance: Healthcare is heavily regulated, with different regions and countries having their own sets of rules and standards. Navigating regulatory compliance while implementing federated learning can be complex and time-consuming.
    \item Bias and fairness: Federated learning may inherit biases present in the data from participating institutions, potentially leading to biased or unfair model outcomes.
    \item Model quality control: Ensuring the quality and consistency of models across different institutions can be challenging. Mechanisms for model monitoring, validation, and quality control are essential to maintain high standards of care.
\end{itemize}
Federated learning is a promising approach for healthcare applications, and addressing the challenges can have significant impacts on healthcare operations.
\subsection*{Crisis management}
\begin{figure}[!hbt]
\centering
\includegraphics[width = 1\textwidth,, keepaspectratio]{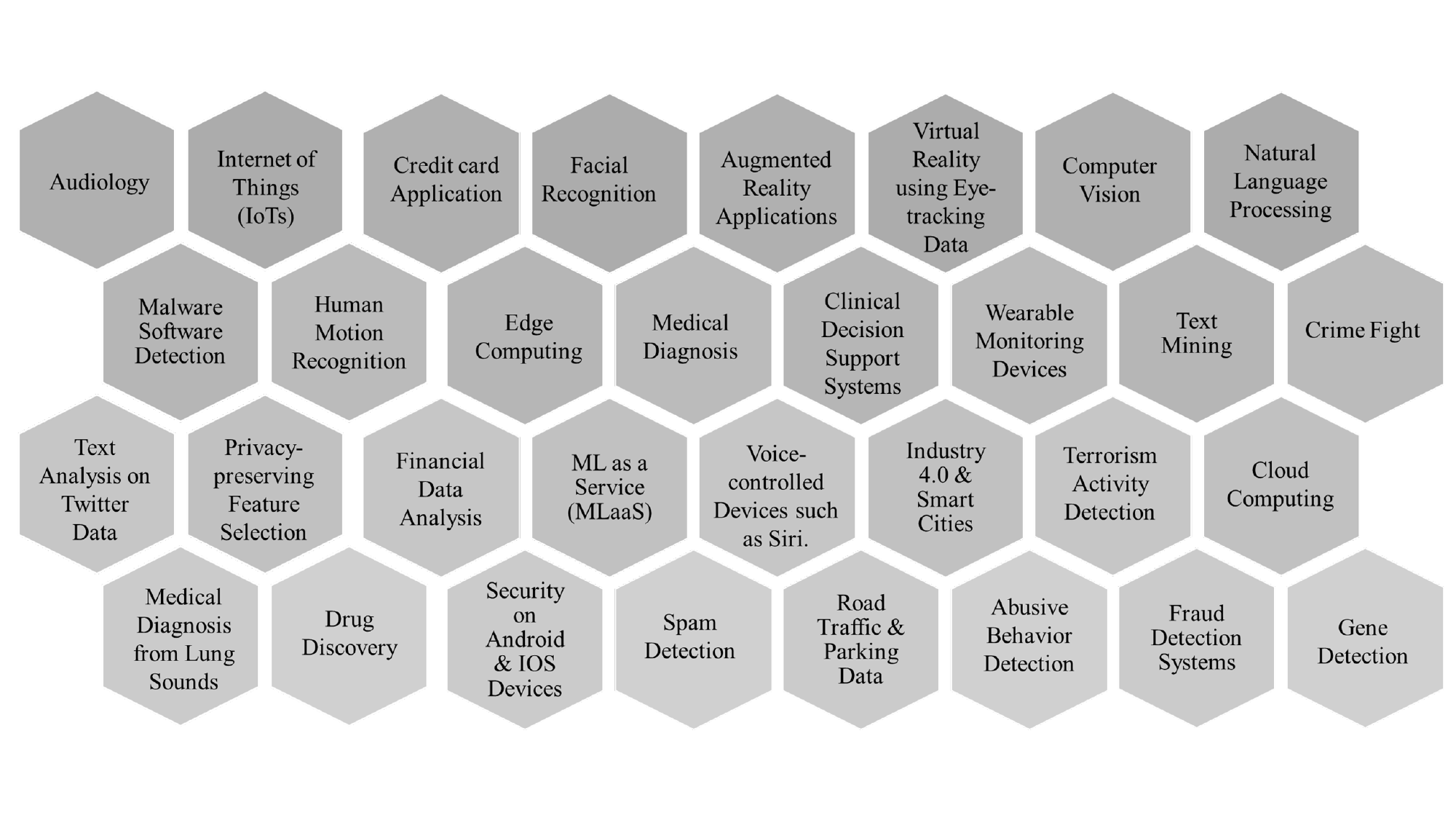}
\caption{Applications of Federated Learning in Industry}
\label{application}
\end{figure}

There is a growing interest in machine learning algorithms for weather applications and natural disasters. Leveraging large data sets from multiple resources facilitates collaborative learning from data collected at different geographic regions, allowing for more powerful and precise models \cite{jafarigol2020imbalanced}. Bypassing the risks of data sharing encourages meteorological institutions across the countries to harness the power of machine learning while ensuring privacy and efficient utilization of resources, resulting in more accurate and timely predictions. On a larger scale, when dealing with natural disasters and emergencies, federated learning allows government agencies, international organizations, local support groups, and communities to collaborate effectively without privacy restrictions caused by traditional centralized learning. Federated learning provides the framework for secure communication of knowledge, resulting in early detection, risk assessment, and effective emergency response strategies using more accurate and robust models.
The benefits of federated learning in crisis management are:
\begin{itemize}
    \item Privacy Preservation: In crisis management scenarios, sensitive and critical data may be involved, such as location data, medical records, or disaster response plans. Federated learning enables multiple entities to collaborate on model training without sharing raw data, preserving privacy and security.
    \item Real-time updates: Crisis management requires quick decision-making based on the latest information.  Federated learning enables real-time model updates as new data becomes available, ensuring that decision support systems remain current and effective during rapidly evolving situations.
    \item Resource efficiency: Crisis response often involves distributed teams and resources. Federated learning leverages the computing power of edge devices and distributed data sources, minimizing the need for centralized data storage and processing resources.
    \item Customization for local conditions: Different regions may have unique characteristics and needs during a crisis. Federated learning allows for localized model customization, ensuring that solutions are tailored to specific conditions and requirements.
\end{itemize}
Despite its benefits, the challenges and limitations of this approach are:
\begin{itemize}
    \item Data availability: Federated learning relies on the availability of data. During a crisis when data may be sparse, incomplete, or unreliable due to infrastructure damage or connectivity issues, training reliable models becomes challenging.
    \item     Heterogeneity: In addition to data availability, collecting data from multiple sources such as governmental and private organization databases, social media, and on-site observations results in inconsistencies in data formats that need special attention during model training and interpretation of results. 
    \item Model drift: In dynamic crisis situations, data distributions can change rapidly, causing model drift which requires the model to be continuously updated and retrained.
\item Communication overhead: Connectivity issues and limitations on bandwidth are possible issues during a crisis. Allocating adequate resources and careful planning prior to emergency scenarios prevents disruptions in critical operations due to disconnections in the network. 
\end{itemize}
Overall, federated learning offers significant potential for enhancing crisis management applications and improving the effectiveness of disaster mitigation and recovery efforts.
\newline 
Figure \ref{application} shows some of the applications of federated learning in the industry based on the papers reviewed in this work.
\subsection{Contributions}
\begin{figure}[!hbt]
\centering
\includegraphics[width = 1\textwidth, keepaspectratio]{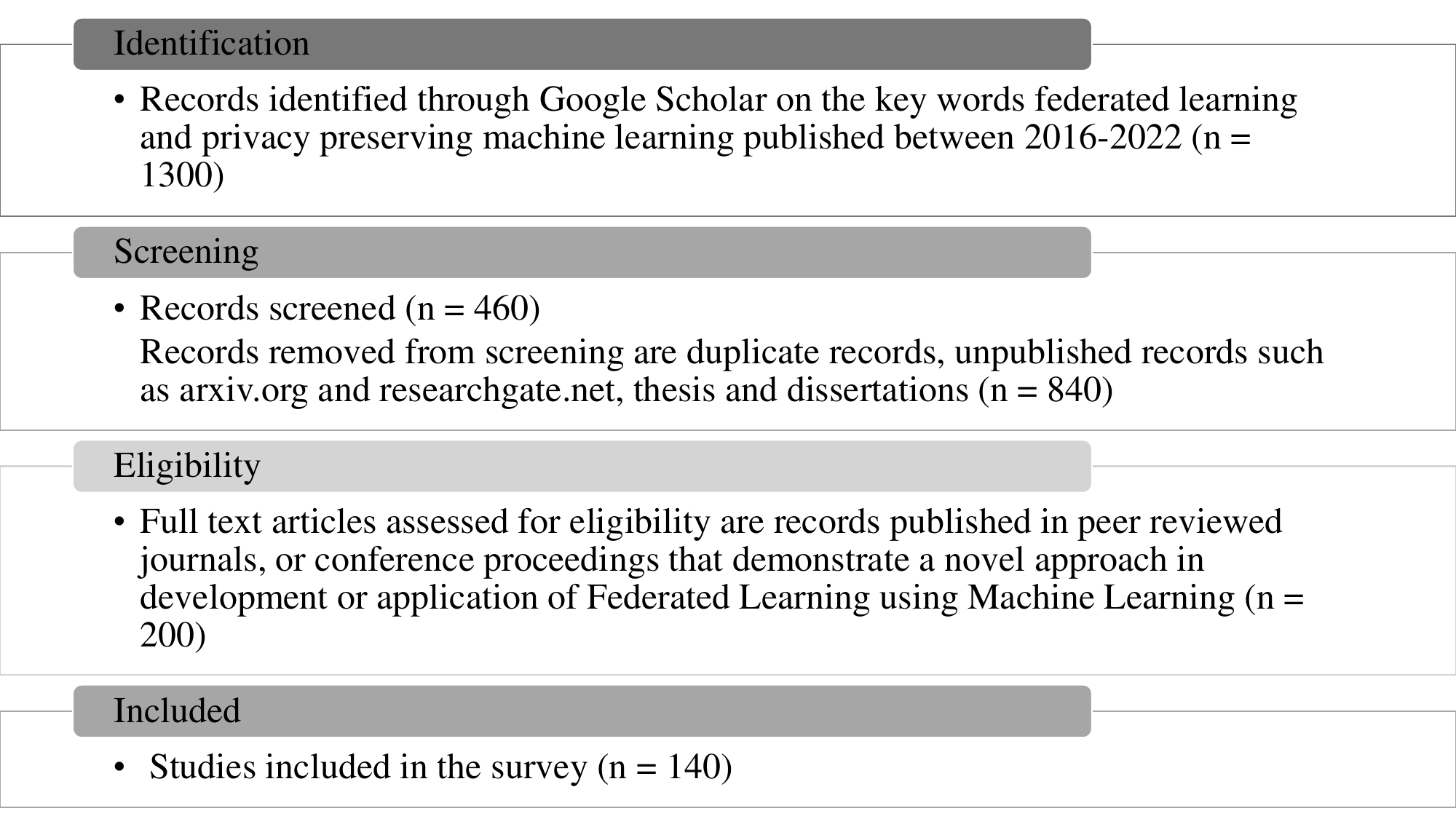}
\caption{PRISMA Flow Diagram Summarizing the Search Strategy }
\label{prisma}
\end{figure}

There is a rich body of research on different aspects of federated learning, such as data processing, learning models, aggregation methods, specifications of data centers and the central server, communication security, and efficiency among the elements \cite{rodriguez2020federated} and the multiple software and libraries used for implementing federated learning\cite{Aledhari2020technology}. In this work, we survey the existing research on federated learning. What distinguishes this work from other surveys is the focus on the model selection aspect of the federated learning process, complementing other recent surveys \cite{ogundokun2022review, gong2020survey, khan2021federated, antunes2022federated, li2021survey, yang2019federated}. We explore the different machine learning models used in federated learning to tackle problems in different domains. In this survey, we have investigated the papers published between 2016-2022 in accredited peer-reviewed journals and conferences and classified them based on the machine learning methods used for learning. We limited the search to the keywords federated learning, privacy-preserving machine learning, distributed learning, supervised/unsupervised learning, and artificial intelligence. The PRISMA diagram presented in Figure \ref{prisma} demonstrates the searching strategy in this survey. 
\section{Components of Federated Learning}\label{section-fl}
In the following sections, we will provide a detailed explanation of the federated learning framework's components: storage, privacy, communication, federated aggregation, and privacy-preserving machine learning, respectively. We dive into a detailed discussion of different machine learning models used for training decentralized data, their use cases and applications, and some technical details useful for implementation. 
\subsection{Storage}
Federated learning is a cross-organizational framework. Therefore, the features and observations may vary between different data centers. Depending on the architecture of the data centers and how the data are partitioned, three scenarios of horizontal partitioning, vertical partitioning, and transfer learning have been discussed. 
\subsubsection*{Horizontal partitioning} Horizontal federated learning, also known as sample-based federated learning, is the scenario in which the data centers have the same features but different sample spaces that require modifications to the training model\cite{kantarcioglu2003privacy}. For example, a network of local banks that individually collect a certain list of information from their clients. In this example, the clients are different. Therefore, the sample space varies between the banks. Horizontal federated learning allows entities to build a generalized global model based on a larger data pool without compromising privacy. 
\newline
\subsubsection*{Vertical partitioning} Vertical federated learning, also known as feature-based federated learning, is when local datasets have the same sample space but differ in their features. Numerous researchers have explored training models specified for vertically partitioned data. For example, a local bank and an insurance company share the same clients but collect different types of information. If the two entities were to build a local model collaboratively, the two datasets would have common sample space but very different features. The common aggregating methods aren't effective in vertically partitioned data. Therefore, the difference between the feature spaces causes different challenges and creates opportunities for further research to address the issue of fault diagnosis.
\newline
\subsubsection*{Transfer Learning} Transfer Learning is the learning structure in which the local datasets differ both in the sample and feature space. Thus, knowledge is derived from various sources to achieve a global model.  Despite all the challenges, transfer learning has been applied to a wide variety of problems in different domains \cite{chen2020privacy}, and it has great potential for further improvement.
\subsection{Privacy}
Preserving privacy is an essential constraint when learning from sensitive data that requires protection against data leakage and adversarial attacks. 
In cybersecurity, the adversary is defined as a person or a group that performs malicious actions to disrupt or corrupt a cyber system. Mothukuri et al. \cite{mothukuri2021survey} identified the different types of adversaries in the federated learning framework and highlighted the solutions to improve systems vulnerability.
Table \ref{adversary} describes the two types of adversaries caused by adversaries.
\begin{table}[!h]
\caption{Types of Adversaries in Cybersecurity} 
\centering
\begin{tabular}{p{2cm}p{1cm} p{5.2cm} p{1.8cm}} 
\hline
Adversaries & Action & Aim & Protocol \\ [0.5ex] 
\hline\hline
Semi-honest & Passive & Learning about the system & \checkmark\\
\hline
Malicious & Active &  Manipulation, corruption, control the system & \xmark\\
\hline
\end{tabular}
\label{adversary}
\end{table}
\newline
To address the problem of adversaries, Secure Multiparty Computation(SMC) is developed. SMC is a framework that enables multiple parties to securely process data while ensuring that no valuable information is leaked. This scheme allows machine learning models to train sensitive data when most of the parties involved are honest. The components are input parties, computation parties, and result parties. For example, SMC is used as a privacy-preserving method in studies on genome data \cite{bogdanov2018implementation}, where the biobanks act as the input parties that hold gene data and medical diagnoses information. The neural hosts and other biobanks are chosen as computation parties, and designated recipients are selected as result parties. This method allows us to combine datasets from genome data collected from different individuals without compromising their privacy. There are different protocols for implementing an SMC that differ in size and number of supported input parties. In highly regulated industries such as healthcare, more advanced privacy-preserving methods are needed to facilitate the use of effective machine learning models in the Federated learning framework \cite{abawajy2017federated}. The privacy-preserving methods are categorized into the two following approaches \cite{lisin2020order}:
\subsubsection*{Perturbation Approach} 
This is a privacy-preserving approach based on adding noise to the data\cite{salim2022perturbation} such as Differential Privacy, which is a perturbation mechanism that involves adding noise to the data to obscure sensitive data and ensuring security in the federated learning framework. Perturbation methods are effective in securing the data, yet decrease the learning accuracy \cite{gursoy2016privacy}. Differential privacy is based on the concept of semantic security, which means the encryption systems prevent enclosing any amount of information through the learning process. In Differential Privacy, the outcome is assumed to be the same regardless of the data if the model is implemented on two neighboring databases. Two datasets of D and D' are considered neighboring if they differ at most in one random variable, and they are written as D$\backsim$D'. This ensures that the algorithm's output does not reveal any information about data sharing and analysis. This feature is also known as $\epsilon$-$DP$. To satisfy Differential Privacy, the Laplace or Gaussian mechanism is used for data with integer or real-valued outputs, and noise is sampled from a Laplace or Gaussian distribution to ensure $\epsilon$-DP for noisy data. The exponential mechanism is used for categorical data, in which each output is associated with a non-zero value for the probability of being selected based on a utility function \cite{truex2017privacy}. First, we compute the sensitivity of the utility function and then compute the quality score of each output in the database. The output is selected probabilistically. Tuning $\epsilon$ in the Differentially Private mechanisms to ensure semantic privacy is one of the challenges of using Differential Privacy for data privacy. Wei et al. \cite{wei2020dp} demonstrate the trade-off between model convergence and privacy level. They use a client scheduling strategy to improve model convergence while maintaining privacy.
Accumulation of noise can jeopardize the accuracy of results if it exceeds a threshold on several operations, and the threshold is on the depth of the operations rather than the number of operations performed to infuse noise to encrypt the data. The depth of the operations is the maximum degree of the evaluated polynomial. The operation depth is determined by the privacy protection scheme as well as the level of speed and security. 
\subsubsection*{Cryptography Approach}
This approach preserves data privacy using cryptographic primitives and includes homomorphic encryption, garbled circuits, secure processors, and order-preserving encryption.  
In this section, we will look closely at Homomorphic Encryption, which preserves privacy and accuracy for the cost of higher running time. The components of Homomorphic Encryption are key generation, encryption, decryption, and evaluation algorithm \cite{dowlin2017manual}. Homomorphic Encryption is a cryptosystem that involves ciphering data using a public or private key and sharing the key among peers to decipher the ciphertext.  Data is ciphered by mathematically transforming the data using addition and multiplication operators \cite{martins2017survey}. Variations of Homomorphic Encryption ensure data security among data centers and the central server. Qin et al. \cite{qin2018privacy} use Homomorphic Encryption for cloud-based privacy-preserving image processing, including feature detection, digital watermarking, and content-based image search.
Limitations of Homomorphic Encryption are that the target space is limited to {0,1} binary values, which is not a feasible representative in practice. There are solutions to address this issue that expand the message space to integers. However, in statistical learning, the values are not limited to binary and integer values. Also, The encrypted ciphertext increases drastically in size, sometimes by several orders of magnitude. This requires additional storage and computation power since the learning procedure is more computationally complex. Current Homomorphic Encryption schemes use only addition and multiplication operators. Therefore, comparison tasks are not supported. Improving encryption schemes to support subtraction and comparison operations, such as inequalities, is an open research question in Homomorphic Encryption. 
\subsection{Communication}
Communication between local data centers and the central server requires sufficient connectivity and bandwidth to ensure secure and private communication between the entities and the entire system. While communication efficiency is highly dependent on the existing infrastructure, reducing the number of interactions between the data centers and the central server can improve the efficiency of the federated learning model. For example, Shen et al. \cite{shen2019privacy} built a blockchain-based model for secure data sharing and training using privacy-preserving Support Vector Machines. Their proposed model requires only two interactions in each iteration, which provides higher performance accuracy and data privacy with less computation cost. In addition, the issue of fairness between the data centers and the central server arises with the federated learning framework. The limitations posed by communication efficiency and security are an ongoing challenge in implementing Federated learning at scale \cite{zhou2020pirate}. 
\subsection{Federated Aggregation}
Secure aggregation is the function that receives model parameters from local data centers and outputs the aggregated model parameters to update the training model. Numerous studies explore aggregation methods to improve learning accuracy in encrypted data. Lia and Togan \cite{Lia2020Secure} implemented federated learning with secure aggregation in Python. The authors ensure privacy by using SMC. 
Federated Averaging(FedAvg) is the baseline aggregating method in federated learning. In this scheme, an initial global model is used to locally train the datasets located on a network of distributed data centers. The encrypted model parameters are uploaded to the central server, and the average updates of local models are used to update and improve the global model. The model parameters provided by the clients are aggregated at the central server using Equation \ref{eq-update}
\begin{equation}\label{eq-update}
\delta_{t+1} \leftarrow \sum_{k=1}^{K} \frac{n_k}{n}\delta^k_{t+1}   
\end{equation}
Then, the new model is sent back to the clients. This iterative process continues until the model parameters converge to a specific performance level or the task is completed. FedAvg is a practical approach since it does not require data centers to disclose their data; thus, the models can be trained locally. However, the communication cost can be high. To address this issue, Li et al. \cite{li2022afedavg} explore an adaptive communication frequency aggregation method that helps the algorithm converge faster and have a smaller loss. They also used a gradient sparse approach to reduce communication costs by decreasing the parameters that need to be updated. Another variation of FedAvg is a weighted FedAvg method, which has been proposed and has shown promising performance in experiments on fault diagnosis \cite{chen2022weighted}. Also, Hong et al. \cite{hong2022weighted} used weighted FedAvg for non-IID imbalanced data and evaluated the model on CIFAR-10 and SVHN benchmark datasets. 
Co-operative aggregation is another aggregation method (CO-OP) in which the local models are merged into the global model by using a weighted scheme based on the local models' age to anticipate the time difference between them and how they have improved at each iteration.
In recent work, Chen et al. \cite{9789131} utilized a discrepancy-based weighted federated averaging method to address the inconsistencies in the contributions of data centers in the global model for federated averaging. The experiments show the effectiveness of federated transfer learning with the proposed averaging method for fault diagnosis.
\subsection{Privacy-preserving Machine Learning}\label{section-ML}
In federated learning, a learning model is built and tuned collaboratively between the central server and data centers \cite{he2014survey}. Chandiramani et al. \cite{chandiramani2019performance} compared the efficiency of numerous machine learning models in the federated learning framework on the benchmark fashion-MNIST data. Learning from distributed data poses different issues and challenges. Therefore, different machine learning models have been explored to compare the performance and efficiency of the learning models. For example, to increase the security in Android devices, Galvez et al. \cite{galvez2021less} built a federated learning malware classification model using K-Nearest Neighbor, Logistic Regression, Random Forest, and Support Vector Machines as machine learning models. 
\begin{figure}[bt]
\centering
\includegraphics[width = 1\textwidth, keepaspectratio]{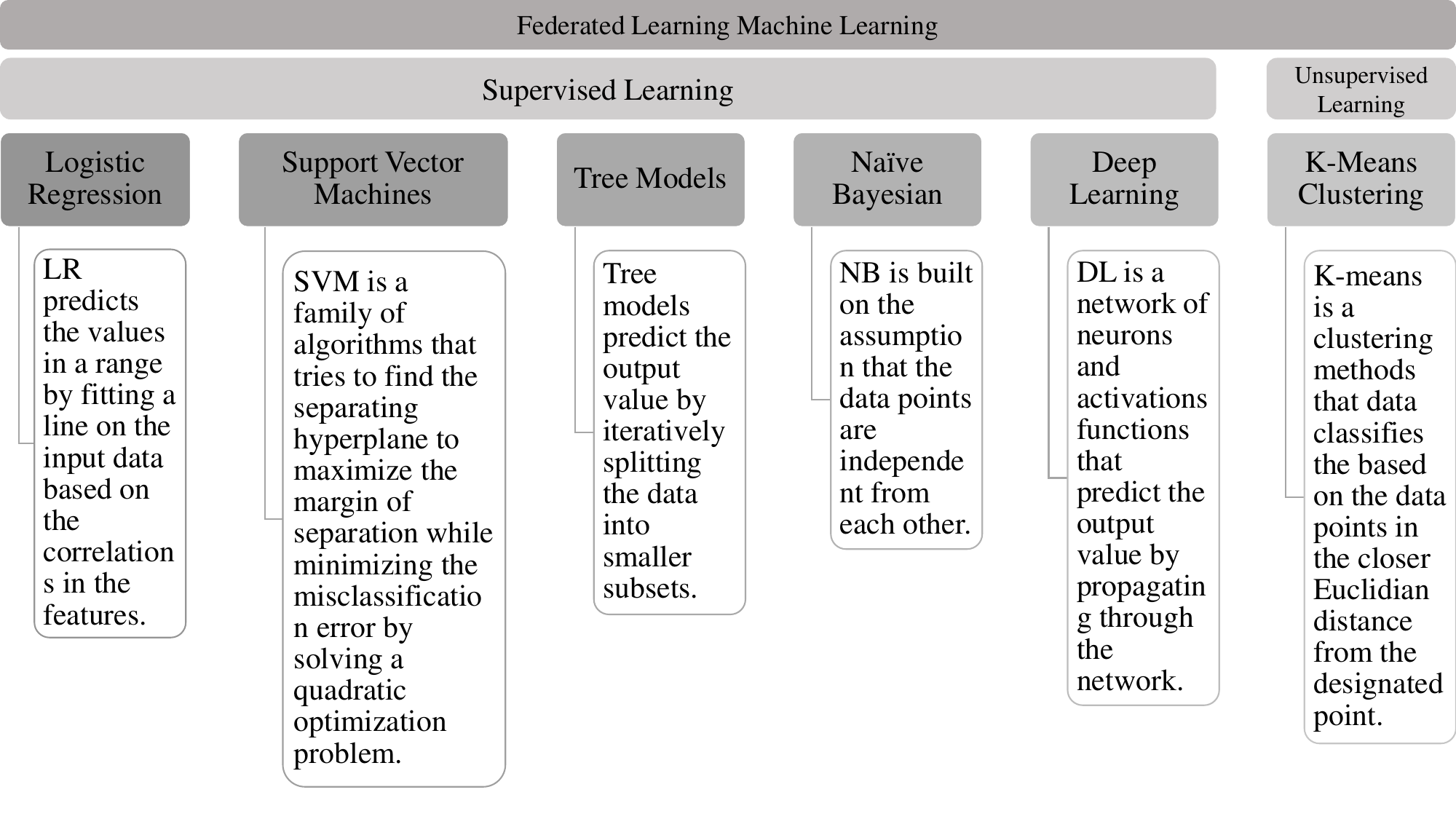}
\caption{Federated Machine Learning Algorithms}
\label{ML}
\end{figure}
\newline
Figure \ref{ML} provides a quick overview of machine learning models used in federated learning literature. 
\newline 
In this section, we have provided a detailed survey of the traditional machine learning algorithms and more recent learning schemes in this domain.
\subsubsection*{Regression Models}
Regression is a predictive modeling approach for identifying the linear and nonlinear relationships between independent variables and the target. Logistic regression has been used in the framework of federated learning for different applications. Yang et al. \cite{yang2020privacy} explore a logistic regression model on clients' credit card and healthcare data. Guo et al. \cite{guo2018privacy} use a logistic regression model to classify illness/health in the cloud environment named POMP. A preprocessing technique and a Bloom filter are also used to reduce the computational complexity in the pre-diagnosis process. The model is implemented using Java and JPBC library. Dunner et al. \cite{dunner2017understanding} implement a ridge regression model on five benchmark datasets to compare the performance of two on Spark and Open MPI, two distributed machine learning frameworks, and suggest recommendations on improving the models implemented in Spark. The results from this paper show that fine-tuning the parameters in a distributed machine learning model to adapt the system's specifications and offloading the Spark language-dependent overheads using C++ can improve computation efficiency. The model is trained for when at least one of the data centers is honest or honest but curious using SPINDLE, an operational system for generalized linear models in distributed learning.\newline
Regression models offer numerous benefits, such as privacy preservation, continuous learning for real-time data analysis, and resource efficiency. However, they deal with the challenges related to heterogeneous data, model aggregation, and privacy concerns, common in many machine learning problems.

\subsubsection*{Support Vector Machines}
Support Vector Machines are widely used in medical diagnosis, spam detection, facial recognition, and analyzing financial data \cite{maekawa2018privacy}. Bost et al. \cite{bost2015machine} use Support Vector Machines to build efficient privacy-preserving algorithms and evaluate the results on breast cancer diagnosis, credit card approval, audiology, and nursery data. Xu et al. \cite{xu2015privacy} implement linear and nonlinear Support Vector Machines on horizontally partitioned data using the MapReduce framework to preserve privacy. Saerom et al. \cite{park2020he} propose a Homomorphic Encryption-friendly least-squares Support Vector Machines to train toy and real-world datasets that outperformed the logistic regression model. Senekane \cite{senekane2019differentially} proposed a privacy-preserving Support Vector Machines framework for image classification. Liang et al. \cite{liang2019efficient} focus on an outsourcing scheme for Support Vector Machines classification with an efficient cryptographic primitive named order-preserving encryption. Chanyaswad et al. \cite{chanyaswad2017compressive} proposed a multi-kernel method using the lossy-encoding scheme to protect the privacy of the data. The training models are Support Vector Machines with an RBF kernel with multiple gamma values and a Signal-to-Noise Ratio-based Support Vector Machines, which is a Signal-to-Noise Ratio for kernel weight design that uses different kernels. The kernel functions are linear, polynomial, Radial Basic Function(RBF), Laplacian, and sigmoid. In terms of privacy, compressing single kernels to form a multi-kernel provides effective results and maximizes utility. The method based on the Signal-to-Noise Ratio method improves the performance compared to uniform and alignment-based methods. Hsu et al. \cite{hsu2020privacy} design a privacy-preserving system for malware software detection using SGD-based Support Vector Machines and SMC techniques. The paper by Zhang et al. \cite{zhang2019human} proposes a solution for the problem of human motion recognition in multimedia interaction scenarios in a virtual reality environment using Support Vector Machines. Chen et al. \cite{chen2020federated} focus on the issue of computation efficiency and low latency of edge computing for augmented reality applications. Hartmann et al. \cite{hartmann2020privacy} proposed a Support Vector Machines model in a privacy-preserving setting, called secret Support Vector Machines, for predicting user gender from tweets based on the online and offline evaluation. Lu et al. \cite{Lu2018multiclass} propose a privacy-preserving feature selection using a multi-class Support Vector Machines named PPM2C, which is evaluated with PAN-SVM and LIB-SVM. The results from this study show that multi-class Support Vector Machines (PPM2C) reduce the chances of overfitting compared to regular Support Vector Machines. Despite being an effective learning method, implementing privacy-preserving Support Vector Machines on data with missing values poses numerous challenges that must be addressed. Omer et al. \cite{omer2017privacy} built a distributed Support Vector Machines model with multiple imputations by chained equations on vertically partitioned data. In this work, the privacy of the data is ensured with the Paillier cryptosystem. The evaluation of the proposed scheme shows higher accuracy and lower computation time compared to the centralized model on imputed data. 
Medical diagnosis systems can significantly benefit from advances in federated learning. Machine learning methods were used to design a secure framework to prevent severe health conditions by diagnosing patients based on their symptoms and the data collected from wearable monitoring devices that monitor heart rate, temperature, oxygen saturation, and other vital signs, and voice-controlled devices such as Google Assistant, Amazon Echo, and Apple Siri \cite{aloufi2020privacy}. Privacy-preserving Support Vector Machines models have been particularly successful in healthcare applications. Wang et al. \cite{wang2020efficient} focus on outsourced Support Vector Machines and EPoSVM (Efficient and Privacy-preserving Outsourced Support Vector Machines) for data classification in the Internet of Medical Things, which results in an improvement in learning accuracy and security compared to Support Vector Machines. Zhu et al. \cite{zhu2016efficient} explored an efficient and privacy-preserving online medical pre-diagnosis framework (eDiag) using nonlinear kernel Support Vector Machines. Ahmed et al. \cite{ahmed2019mlung} developed "mLung", a cloud-based privacy-preserving service to detect chronic pulmonary issues from lung sounds such as cough. The analysis is performed on a personal mobile phone to ensure privacy. Medical components of the drugs must be kept private by pharmaceutical companies. Wang et al. \cite{Wang2018collaborative} explored an encrypted kernel Support Vector Machines using Homomorphic Encryption. They have built a sub-image to position different sections of the image so that a face can appear. The authors trained and tested their model on the BioID Face Database. \newline 
One of the key benefits of Support Vector Machines in federated learning is their ability to generalize well from limited data, making them suitable for scenarios where each participant has a relatively small dataset. However, some of the challenges and limitations of Support Vector Machines are:
\begin{itemize}
    \item Complexity of kernel functions: Support Vector Machines rely on kernel functions to handle the data in nonlinear space. However, selecting appropriate kernel functions in a framework with diverse data sources can be challenging, as different participants may require different kernel types.
    \item Communication overhead: Support Vector Machines are computationally expensive due to dealing with large datasets and complex kernel functions, making them unsuitable for edge devices or environments with limited computational power.
    \item Hyperparameter tuning: Support Vector Machines have multiple hyperparameters, such as the regularization parameter (C) and the kernel parameters. Hyperparameter tuning across multiple participants in a federated setting can be complex and time-consuming.
\end{itemize}
While Support Vector Machines can be computationally expensive, they offer strong generalizations that can lead to robust models even with limited local data.

\subsubsection*{Tree Models}
Decision Tree and Random Forest are a group of effective and widely used models for data classification and regression \cite{li2017random}. Khodaparast et al. \cite{khodaparast2018privacy} presented a Decision Tree algorithm equipped with a federated learning framework for horizontal and vertically partitioned data. Yadav et al. \cite{yadav2016privacy} presented a new Decision Tree-based model in a privacy-preserving manner in which the data are partitioned vertically into multiple parties, and the parameters are sent to the central server. Badsha et al. \cite{badsha2019privacy} present a privacy-preserving Decision Tree framework to build and learn the tree-based model without requiring the parties to disclose private information. The authors use Homomorphic Encryption to maintain privacy, while the parties are assumed to be honest but curious. The Gini Index is used to measure the classification capability of the model. Canillas et al. \cite{canillas2018exploratory} explored a privacy-preserving Decision Tree framework for private fraud detection systems at SiS ID, a French business platform. The model is used to classify transactions into four risk classes. The model's accuracy depends on the configuration of the encryption key and the number of nodes for the Decision Tree. The proposed model utilizes a Decision Tree algorithm that helps improve the diagnosis pace and accuracy based on the patient's symptoms without disclosing the patient's private data. Xue et al. \cite{xue2020consent} propose a consent-based privacy-preserving Decision Tree model for the evaluation scheme. The additive Homomorphic Encryption method and a secure comparison model are used. Also, Xue et al. \cite{xue2020secure} propose a privacy-preserving Decision Tree for classification using additive Homomorphic Encryption, which provides lower computation and communication overhead. 
Hou et al. \cite{hou2019dprf} explore Random Forest with a Decision Tree for data classification and study the impact of tree depths in Decision Tree on privacy and classification. A privacy budget is allocated for nodes at different depths in the Decision Tree. Guan et al. \cite{guan2020differentially} explore a budget allocation mechanism for Decision Tree construction for balancing the excessive noise introduced at leaf nodes. The iterative process speeds up the selective aggregation process. The tree is constructed based on the C4.5 method. Lv et al. \cite{lv2019differential} proposed a hybrid Decision Tree algorithm for constructing a Random Forest to balance privacy and classification accuracy. In the paper by Xin et al. \cite{xin2019differentially}, the authors propose a new differentially private greedy Decision Tree algorithm called (DPGDF) which is a combination of greedy trees and parallel combination theory. Zhao et al. \cite{zhao2018inprivate} explore a tree-based data mining model for regression and binary classification tasks. In this work, a privacy-preserving Gradient Boosting Decision Tree (GBDT) model is aggregated into an ensemble. Random Forest is also used for feature engineering. Fritchman et al. \cite{fritchman2018privacy} propose a tree ensemble approach to learn from the data collected in healthcare institutes securely. Zhang et al. \cite{zhang2019pdvocal} has built a secure Parkinson's diagnosis framework using non-speech body sounds such as breathing and coughing.\newline
The main advantages of tree-based models, including Decision Trees and Random Forests, are their interpretability, ensemble learning for robustness, and feature importance analysis.

\subsubsection*{Naïve Bayesian Algorithms}
Naïve Bayesian algorithms is a supervised learning algorithm centered around the Bayes theorem, which is based on the assumption that there is conditional independence among each pair of features if the class is known. In smart environments such as smart cities, data privacy is crucial. Amma and Dhanaseelan \cite{amma2018privacy} explore a Naïve Bayesian classification framework for privacy-preserving machine learning on the cloud using smart city data. The authors validated the results using Viz road traffic, pollution, and parking data collected from the City Pulse Smart City dataset. Part et al. \cite{parto2020novel} introduced a novel federated learning architecture that consists of three layers. In the edge layer, the data is processed, the machine learning models are trained in the fog layer, and the results are aggregated in the centralized cloud layer. Data partitioning poses specific challenges when learning from the data in a distributed setting. Vaidya et al. \cite{vaidya2008privacy} investigate a privacy-preserving Naïve Bayesian classification model and compare the results on horizontal and vertically partitioned data. Yurochkin et al. \cite{yurochkin2019bayesian} propose a Bayesian non-parametric federated learning framework with neural networks. The model is evaluated on two benchmark data sets for image classification. The paper by García-Recuero \cite{garcia2016discouraging} addresses the issue of detecting and discouraging abusive behavior in online social networking applications such as Twitter by limiting the accessibility to user-sensitive information. This work uses feature engineering on relative importance calculated from the Random Forest learning algorithm. Li et al. \cite{li2018outsourced} use Naïve Bayesian and a hyperplane-based decision model for classification. Furthering this work, Chai et al. \cite{chai2020improvement} propose an outsourced encryption protocol to improve the security vulnerability of Li's model. The classification models are trained using Scikit-learn. Paillier cryptosystems are used in this work due to light operations. In a paper by Yang et al. \cite{yang2018efficient}, the authors propose a communication-efficient privacy-preserving framework based on the Naïve Bayesian method to predict the disease risk for e-health applications.
Sharkala et al. \cite{skarkala2020pp} introduce a privacy-preserving machine learning algorithm for horizontally and vertically partitioned data based on a tree augmented Naïve Bayesian classifier. A third party conducts the operations, and the data is encrypted. Teo et al. \cite{teo2018dag} propose a privacy-preserving algorithm using kernel regression and Naïve Bayesian classifier for multiparty computation, and the Paillier cryptosystem is used for encryption. Medical diagnosis systems are the main application of privacy-preserving machine learning. Liu et al. \cite{liu2018efficient} build a secure diagnosis scheme using a Naïve Bayesian classifier. Furthermore, malware detection systems protect the user's identity by identifying malware API call fragments. Lin et al. \cite{lin2018secure} propose a privacy-preserving Naïve Bayesian model for malware detection. Talbi et al. \cite{talbi2018towards} compare their classification algorithms of Naïve Bayesian, Decision Tree, and logistic regression on encrypted data. \newline
Naive Bayesian algorithms provide probabilistic predictions and can quantify uncertainty. In federated settings, this is a significant advantage for risk assessment and decision-making, particularly in applications such as crisis management and disaster response scenarios where uncertainty plays a significant role.
 
\subsubsection*{Deep Learning}
Deep learning has been the dominant learning method from structured and unstructured data in recent years. Deep learning is the burgeoning powerful technique in the field of machine learning. Deep architectures are useful for learning complicated patterns in large-scale data, attracting much attention in academia and industry. Different topologies and architectures with real-world applications exist. These models have been used in different areas such as Computer Vision, Natural Language Processing, and speech recognition \cite{zhang2018verifiable, he2020transnet}. The improvements in implementing a secure Neural Network model on the cloud create a platform for scalable Neural Networks to be used as a service \cite{hesamifard2017privacy, hesamifard2018privacy}. A Recurrent Neural Network is a variation of a forwarding propagation Neural Network in which the neurons in the hidden layers receive the input value with a delay in time and access information from previous iterations in the current layer. Recurrent Neural Network is useful in Natural Language Processing, where knowledge about the previous words in a sentence is necessary for predicting the next word. Text mining and Natural Language Processing, which is the process of extracting knowledge from text documents, are used for learning from data collected from highly sensitive resources such as homeland security for crime fight and detecting terrorism activities. Therefore, building secure federated learning text analysis methods is necessary to ensure no sensitive information is disclosed. To this end, Costantino et al. \cite{costantino2017privacy} propose using an out-of-bag classification method to detect terrorist activities on Twitter. Convolutional Neural Network is a deep learning architecture that has gained popularity in Computer Vision \cite{rahim2018privacy}. This architecture uses one or more convolutional layers to extract high-level features. Wang et al. \cite{wang2018privacy} investigate a privacy-preserving Natural Language Model framework to compute word representations using deep learning. Abadi et al. \cite{abadi2016deep} explore training a Neural Network model with a non-convex objective function, implemented using TensorFlow in Python, with differentially private Stochastic Gradient Descent method, the moments' accountant, and hyperparameter tuning. The security protocol is a semi-honest security model that is a secure computation method based on additive secret-sharing techniques (secure addition, subtraction, and secure multiplication). Xia et al. \cite{xia2022cnn} designed a Graph Convolutional Network model for predicting traffic flow quickly and efficiently.
Deep learning methods can also be combined with other classifiers, such as linear Support Vector Machines, for the classification of images.  Niu et al. \cite{niu2018privacy} explore a deep learning framework for mobile sensing systems. Lin et al.\cite{lin2017pcd} propose a Recurrent Neural Network framework called the Predictive Clinical Decision (PCD) scheme, which is used for e-health applications.
Eye-tracking devices are the main technology in virtual reality and augmented reality that can improve efficiency through gaze-based optimization methods. The eye-wear and eye-tracking devices used in the auto industry can pose privacy issues for the driver and bystanders \cite{bozkir2019person}. Therefore, Steil et al. \cite{steil2019privaceye} explore a privacy-preserving method for a first-person video dataset of daily life recordings. The authors propose the PrivacEye method that combines Computer Vision with eye movement analysis techniques. Another privacy concern with Computer Vision technologies in facial recognition systems such as Google Street View is when personal images of individuals are shared via different data centers. \newline 
Deep learning models have demonstrated great potential for highly accurate and competitive results when dealing with diverse and large datasets. While increasing model complexity in neural network architectures helps improve generalization, excess complexity results in overfitting and computation overhead. Adaptation, optimization, and the integration of privacy-preserving techniques are essential to harness the strengths of deep learning while mitigating the specific challenges and limitations present in federated settings.

\subsubsection*{Unsupervised Machine Learning}
Current Federated learning machine learning-based models are constructed based on supervised learning. However, in most applications, no or little labeled data exists. Thus, it is appropriate to use unsupervised learning methods. While there has been considerable progress on federated transfer learning to cope with data with few labels, applying unsupervised learning in a federated setting remains a bottleneck for many applications. Clustering techniques have been employed to deal with the challenges of unlabeled data. While K-means clustering is widely used for pattern recognition in gene detection and image segmentation, a modified framework is required when the data is sensitive. Zhu and Li \cite{zhu2020privacy} proposed a secure aggregation and division protocol based on Homomorphic Encryption to build a secure clustering algorithm. Al-Saeidi et al. \cite{alsaeidi2021clustering} proposed a clustering analysis for improving the communication cost in federated learning using the human activity recognition dataset. A secure weighted average protocol and secure number comparison protocol are used for privacy-preserving. Five different classification algorithms were explored: multi-layer perceptron, K Nearest Neighbor, Sequential Minimal Optimization, Naïve Bayesian, and J48 (an implementation of Decision Tree classification in WEKA). Anikin and Gazimov \cite{anikin2017privacy} proposed a clustering algorithm named Density-Based Spatial Clustering of Applications with NOISE (DBSCAN) for vertically partitioned data. Romsaiyud et al. \cite{romsaiyud2019improving} investigate a privacy-preserving K Nearest Neighbor model for pattern recognition, with automated hyperparameter tuning to improve model accuracy and a cryptographic hash function to ensure data security. \newline 
The benefits of unsupervised learning techniques in federated learning are:
\begin{itemize}
    \item Privacy-preserving clustering: Clustering algorithms can perform data analysis without the need for explicit labels or excess information sharing, which is beneficial when dealing with sensitive data.
    \item Data exploration and anomaly detection: Unsupervised models excel at data exploration, allowing practitioners to identify the underlying patterns, anomalies, and outliers within local datasets. This exploratory capability is valuable for uncovering insights without exposing private data.
    \item Reduced labeling effort: Limited labeled data is a well-known issue in machine learning. Unsupervised learning models can reduce the labeling effort by enabling semi-supervised or self-supervised learning approaches without exposing the data.
\end{itemize}
With appropriate evaluation strategies, we can effectively leverage the strengths of unsupervised learning while mitigating the specific challenges and limitations of federated learning.

\subsubsection*{Ensemble Learning}
Ensemble learning is a general approach that seeks to improve learning performance by aggregating the results from multiple classifiers. The paper published by Attota et al. \cite{Attota2021ensemble} used an ensemble multi-view federated learning model to identify intrusion in IoT devices to improve model efficiency against different attacks. Ma et al. \cite{ma2019lightweight} apply edge computing methods for medical diagnosis using the XGBoost model. They use a lightweight, adaptive boosting classification method (AdaBoost) for facial recognition on FERET, a standard face recognition evaluation database. The data is encrypted when sent to two servers for distributed learning. \newline 
Ensemble learning is an effective approach in the machine learning domain and has yet to be extensively explored in the federated learning framework. There is a very limited number of papers published in peer-reviewed journals using this approach. However, they demonstrate the potential of ensemble learning in federated learning. 

\subsubsection*{Meta-heuristic Approaches}
Apart from the more commonly used machine learning algorithms mentioned above, meta-heuristic approaches have also been introduced to the federated learning domain. Polap and Wozniak \cite{polap2022hybridization} used a novel approach based on parallelism to improve classification models' efficiency in federated learning. The authors demonstrate the effectiveness of their approach when the sample size is relatively small. In another paper, \cite{polap2021meta}, they explore using a meta-heuristic federated learning framework for image classification in the presence of poisoning attacks. With application in IoT and smart city services, Qolomany et al. \cite{qolomany2020particle} investigate using Particle Swarm Optimization for efficiently tuning the hyperparameters in the machine learning model. Utilizing meta-heuristic approaches can be further explored as a novel approach to improve the efficiency of the training model in federated learning. \newline 
Metaheuristic algorithms are versatile and adaptable to various problem domains. They can be customized to suit the specific requirements and constraints of federated learning scenarios.

\subsubsection*{Blockchain Technology}
Despite improvements in Homomorphic Encryption and Differential Privacy in preserving privacy, there is always a trade-off between learning accuracy and privacy. To overcome such issues, federated learning can be equipped with blockchain technology \cite{chen2022blockchain, ali2021integration}. Utilizing blockchain technology in federated learning is an emerging field in federated learning and decentralized data storage and processing \cite{Lu2020blockchain, Pokhrel2020blockchain}. Nguyen et al. \cite{Nguyen2021blockchain} survey the advances and challenges of federated learning with blockchain technology. In edge computing and learning from IoT data, blockchain federated learning is a solution to issues in data storage, communication cost, and privacy of sensitive data. Wang et al. \cite{WANG2022, WANG2021BLOCKCHAIN} use a blockchain-distributed setting as the groundwork for federated learning to ensure additional privacy protection from servers and prevent malicious attach on user-sensitive data.  Kang et al. \cite{Kang2020mobile} proposed a federated learning framework based on a blockchain mechanism and introduced reputation as a metric to identify reliable data and propose a reliable framework to learn from the data on mobile networks. Later, Kang et al. \cite{Kang2022} used multiple blockchains to design a cross-chain framework to improve the scalability and communication efficiency of federated learning for training the data on IoT devices. An example of other applications is a classification of COVID-19 cases from multiple resources using blockchain-based federated learning. Comparing the federated learning framework to centralized models shows improvement in diagnosing COVID-19 patients \cite{9420107}. \newline 
Leveraging blockchain technology within the context of federated learning introduces many benefits, such as:
\begin{itemize}
    \item Data privacy and security: Blockchain technology has inherent privacy protection capabilities allowing participants to maintain control over their data while securely contributing to the global model.
    \item Transparent and trustworthy transactions: The decentralized nature of blockchain ensures that all transactions and updates are transparent and traceable. This feature mitigates concerns of data tampering and adversarial attacks.
    \item Smart contracts for governance: Smart contracts are programmable scripts executed on the blockchain that can be employed for governing federated learning agreements and model updates. This automation streamlines the process and enforces predefined rules and policies, reducing the risk of malfunction and misuse.
\end{itemize}
Despite its benefits, implementing a blockchain-based federated learning system is complex and requires expertise in both blockchain technology and machine learning. Developing and maintaining such a system is challenging as some blockchain networks consume a significant amount of energy. This environmental impact may not align with sustainability goals in federated learning.

\subsubsection*{Reinforcement Learning} 
Incorporating other learning approaches into federated learning has shown promising results in different applications \cite{Hyesung2020blockchain}. Liu et al. \cite{liu2019deep} built a reinforcement learning framework, which is a learning system through trial-and-error interactions between agents and environments combined with cloud computing and IoT technology to create a dynamic system for cancer patient treatment regimes. Wang et al. \cite{wang2020reinforcement} proposed a reinforcement learning mechanism to introduce a rewards system that optimizes accuracy and communication efficiency. A reinforcement learning approach is also used for evaluating node contributions and improving the pricing strategy in federated learning for IoT devices \cite{Zhang2020IoT}. Krouka et al. \cite{Krouka2022reinforcement} investigate different aggregation schemes in a reinforcement learning-federated learning framework to improve communication costs.\newline 
The benefits of combining reinforcement learning and federated learning are :
\begin{itemize}
    \item Dynamic model adaptation: Reinforcement Learning models can adapt dynamically to changing data distributions and evolving environments. In federated learning, where data sources may drift or have different characteristics, Reinforcement Learning can facilitate model adjustments for improved performance.
    \item Sequential decision-making: Reinforcement Learning is well-suited for sequential decision-making tasks. In federated learning scenarios, this capability can be valuable for applications that involve sequential interactions or recommendations.
\end{itemize}
Reinforcement Learning models often require extensive training and exploration of different policies, which can be computationally expensive and time-consuming. Reinforcement Learning must efficiently find the optimal policy by balancing the exploration of new actions and the exploitation of known policies. Managing the trade-off between exploration and exploitation is necessary to avoid excessive data sharing or overfitting.

\section{Challenges and Open Problems}\label{section-challenges}
Federated learning is at the interface of several research areas, such as optimization, distributed learning, cryptography, and communication theory. In the last few years, many algorithmic developments have been accomplished with a focus on theory and applications. However, there are still several challenges and open problems in federated learning. In a recent paper, some advances and open problems are discussed \cite{kairouz2021advances}. 
In addition, central to the future of federated learning are the data-driven challenges outlined in Table \ref{tab-issues}.
\begin{table}[!h]
\caption{Data-driven Issues in Federated Learning}
\centering
\begin{tabular}{p{2.5 cm} p{0.1cm} p{7.5cm} p{0cm}}
\hline
Issues &\ &\  Challenges &\\
\hline\hline
Unlabeled Data & & Labeling the data require considerable time and resources.\\
\hline
Non-IID Data & & Different distributions in data centers negatively affect learning performance and efficiency. \\
\hline
Imbalanced Data & & imbalanced Data result in reduced learning performance in the minority class. \\
\hline
\end{tabular}
\label{tab-issues}
\end{table}
\newline
The data-driven issues pose various challenges for efficient and accurate training of federated learning models. 
\begin{itemize}
    \item Unlabeled data: The lack of annotated data with good quality is one of the limitations in this setting. When the data is unlabeled, and labeling the data is either impossible or too cost-inefficient, it is important to modify the machine learning algorithms to be able to learn from partially annotated data efficiently.
    \item Non-Independent and Identically Distributed (Non-IID) Data: Non-IID data in federated learning refer to differences in the distribution of the available data over the data centers.  It is also possible that the data become locally non-IID over time, which requires modifying existing algorithms or developing new ones. To address this issue, Sattler et al. \cite{8889996} proposed a compression network to improve the communication efficiency and robustness of Federated learning on non-IID data. Ma et al. \cite{ma2022state} and Zhu et al. \cite{zhu2021federated} investigated the recent advances in solving non-IID data in federated learning and the future trends in research on this issue. They also compared different model architectures of deep learning used in the literature on federated learning. 
    \item Imbalanced data: The data is imbalanced when the number of instances in one class is significantly larger than in the other. This issue can arise in the taxonomy of the data centers or the overall data. The challenges posed by data-driven issues require experiments on achieving highly efficient and accurate learning systems.
\end{itemize}    
Other challenges that need to be overcome are:    
\begin{itemize}
    \item Fairness: The issue of fairness in federated learning occurs at different levels, such as fairness in communication capacities, machine learning models, and aggregation results. To this end, fairness metrics must be defined to evaluate the model from privacy, accuracy, and fairness perspectives \cite{yu2020fairness}.  Also, Lyu et al. \cite{lyu2020collaborative} propose the concept of collaborative federated learning, which ensures fairness in how the clients impact the global model in the aggregation process. Despite the improvements, there is a lack of an integrated approach that ensures fairness in different aspects of federated learning.
    \item Scalability: To implement a federated learning protocol at scale, it must avoid the curse of dimensionality when data is large. To tackle the challenge of dimensionality, Principal Component Analysis has been employed in unsupervised settings such as the work by Al-Rubaie \cite{al2017privacy}, which developed a privacy-preserving Principal Component Analysis to reduce the dimensionality of the horizontally partitioned data. In addition, other methods, such as Discriminant Component Analysis, can be explored for efficient feature engineering and dimensionality reduction to improve accuracy and computation cost and uphold privacy.
\end{itemize}
Possible future research directions are:
\begin{itemize}
    \item Game theory: Recent connections between game theory and control can provide new insights into federated learning and new algorithmic developments. An attempt along these lines is the paper of Mehrjou\cite{Mehrjou2021FederatedLA}, which connects federated learning with mean field games by presenting federated learning as a differential game and discussing the properties of the equilibrium of this game.
    \item Quantum optimization:  Another future research area is quantum optimization applied to federated learning. Distributed learning across several quantum computers could significantly improve the training time and potentially improve data privacy \cite{chen2021federated}. Connections with multi-objective optimization,  such as \cite{zhong2022optimizing}, can benefit algorithmic developments.
    \item Multiple Kernel Learning: Exploring the connections between federated learning and multiple kernel learning also holds potential for advancing algorithmic solutions.
\end{itemize}
Furthermore, federated learning must navigate complex challenges, including exploring more advanced ways of combining local networks, using different machine learning and ensemble models in addition to preprocessing techniques, performing experiments on larger datasets and further enhancement efficiency, and designing accurate and reliable machine learning models suitable for GPU implementation.
In summary, federated learning stands as a field where innovation and collaboration are essential. By collectively addressing these challenges and pursuing interdisciplinary connections, The research community can integrate federated learning into a future where it serves as a cornerstone for secure, efficient, and privacy-conscious machine learning applications.

\section{Conclusion}\label{section-conclusion}
We have conducted an extensive literature review of federated learning from the machine learning point of view, complementing other recent literature reviews. This review will be useful for researchers in academia and industry and possibly a useful tool for graduate students who want to work in this area. Federated learning was proposed as a solution to the issue of data leakage and loss of privacy in machine learning. With a large amount of data at hand, there is a burgeoning demand for federated learning as potentially being the solution to private and environmental-friendly machine learning at scale. Decentralized learning strategy and privacy protection mechanisms in federated learning grant us access to otherwise unavailable data. Hence, we can expand machine learning in domains such as IoT and healthcare and crisis management in natural and human-caused disasters that require privacy preservation. In recent years, there has been an increasing number of papers published in this domain, and the goal of this study is to provide an overview of federated learning and the existing privacy-preserving machine learning algorithms used in this framework, in addition to their potential and limitations in various applications. Despite our effort to thoroughly search the literature on federated learning, we limited our search to published papers in peer-reviewed journals in English. Therefore, other novel approaches might be found in the papers not included in this survey. 
\backmatter
\section*{Declarations}
\textbf{Conflict of interest} The authors declare no competing interests

\bibliography{bibliography}
\end{document}